# White Paper: Challenges and Considerations for the Creation of a Large Labelled Repository of Online Videos with Questionable Content

Thamar Solorio[1], Mahsa Shafaei[1], Christos Smailis[1], Mona Diab[2], Theodore Giannakopoulos[3], Heng Ji[4], Yang Liu[5], Rada Mihalcea[6], Smaranda Muresan[7], Ioannis Kakadiaris[1]

[1]University of Houston, [2]Facebook, [3]National Center for Scientific Research "Demokritos", [4]University of Illinois Urbana Champaign, [5]Amazon, [6]University of Michigan, [7]Columbia University

## Abstract

This white paper presents a summary of the discussions regarding critical considerations to develop an extensive repository of online videos annotated with labels indicating questionable content. The main discussion points include: 1) the type of appropriate labels that will result in a valuable repository for the larger AI community; 2) how to design the collection and annotation process, as well as the distribution of the corpus to maximize its potential impact; and, 3) what actions we can take to reduce risk of trauma to annotators.

## A. Background

Due to the proliferation of online streaming platforms, users of all ages are turning to online surfaces as a popular entertainment activity. Content in these streaming platforms might often include material that parents/guardians would not like their children to see. Moreover, some of these online videos have been shown by psychologists and mass media experts to negatively affect young viewers (Bridges et al., 2010; Bushman & Anderson, 2009; Chang & Bushman, 2019; Dillon & Bushman, 2017; Gentile et al., 2011, American Academy of Pediatrics, 2001; Hanewinkel et al., 2014; Swim et al., 2003). At the same time, a large number of videos made available daily makes it infeasible to screen videos manually. Therefore, we want to support the development of multimodal technology that can automatically label questionable content in videos. Interested users (adults, as well as parents and/or guardians) can then use these labels to decide if the video is appropriate for them or their children.

While most online streaming service providers include some ability to apply age related or content related filters, these filters have been found to be easily deceived. On the other hand, state of the art approaches have been proposed to detect specific types of questionable content in specific modalities, violence and nudity in videos, and verbal aggression and sarcasm (negative content) in text or speech. What is still missing is technology that can address questionable content in a comprehensive manner. Not only in terms of the type of questionable content, but also with respect to the modality of the content.

To advance this research agenda, the first step is to create a labeled data set to train, tune, and evaluate automated approaches to support the development of such technology. Creating the labeled data set represents a considerable effort, especially since the goal is to create an infrastructure that supports research across multiple AI disciplines.

The lead research team has hosted a series of workshops targeting different participants' expertise to gather relevant feedback to design and create a sizable multimodal repository of online videos labeled with tags indicating the presence of potentially questionable content. This white paper relates to the fourth online workshop that gathered world-known experts in Natural Language Speech/Signal Processing. What follows is an overview of the major points discussed at the workshop.

**B. Annotating a Multimodal Repository with Labels of Questionable Content**

A video is a collection of scenes that may or may not include questionable content. Labeling an entire video as having questionable content may be a good starting point. Still, the most relevant and useful scenario would be to label scenes with tags that indicate the presence and type of questionable content. Our goal will be to segment a video into scenes and label each scene. We will explore the use of existing video content processing tools, including scene parsing, object detection, and language grounding to speed up the process of annotation and/or screening.

<u>Defining Questionable Content</u>

The exact delineation of what represents questionable content is hard to formulate as it is culture-dependent. Instead of aiming for a narrow and culture-specific definition, we can aim for a broad set of carefully defined labels known to be questionable by people from different cultural/religious/demographic backgrounds while at the same time prioritizing perhaps the content that has been documented to create adverse effects in viewers in the relevant literature. Tables 1-4 show the categories of questionable content we are considering for each of the four modalities (dialogue, sound, soundtrack and video) and their definition.

| Label | Definition |
|---|---|
| Mature Humor | Humor involving sexual references in a video dialogue |
| Slapstick Humor | Dialogue references of humor involving mild physical or cartoon violence - Should be viewed as the intent of the video creator to produce comedy using violent elements |
| Gory Humor | Dialogue references of humor involving intense physical violence |
| Sarcasm | Passive-aggressive humorous comments at the expense of another person or situation |
| Fear Horror | A scary story such as talking about hunting, ghost, summon the soul |
| Gambling | Discussions that glamourize gambling |
| Foul Language | Use of profanity in dialogue |
| Hate Speech | Speech that expresses hate or encourages violence towards a person or group based on something such as race, religion, sex, or sexual orientation |
| Sexual Themes | References to sex related content dialogue |

| Alcohol | Dialogue references to alcoholic beverages |
|---------|--------------------------------------------|
| Drug | References to drugs in dialogue |
| Tobacco | Dialogue references the use of tobacco |
| Violence | Dialogues describing violent acts |

Table 1. Labels for questionable content on the dialogue modality

| Label | Definition |
|-------|------------|
| Fear/Horror | Spooky, suspense, horror sound effects or screams |
| Sexual Themes | Sounds related to sexual scenes |
| Violence | Sounds indicating violent acts (e.g., the sound of slashes, the sound of explosions). There is no need for somebody to be physically hurt to use this label. |
| Blood/Gore | Sounds implying gory actions (e.g., dismemberment) |

Table 2. Labels for questionable content in the sound modality

| Label Name | When to select |
|------------|----------------|
| Mature Humor | References in soundtrack lyrics to humor involving sexual references |
| Slapstick Humor | References in soundtrack lyrics to humor involving mild physical or cartoon violence (should be viewed as the intent of the video creator to produce comedy using violent elements) |
| Gory Humor | References in soundtrack lyrics to humor involving intense physical violence |
| Sarcasm | References in soundtrack lyrics to passive-aggressive humorous content at the expense of another person or situation |
| Fear/Horror | Spooky, suspense, horror Music |
| Gambling | Lyrics that glamourize gambling |
| Foul Language | Use of profanity in music lyrics |
| Hate Speech | Use of hate speech in music lyrics (hate speech is defined as speech that expresses hate or encourages violence towards a person or group based on something such as race, religion, sex, or sexual orientation) |
| Sexual Themes | References to sexuality in soundtrack lyrics |
| Violence | Soundtrack lyrics describing violent acts |

Table 3. Labels for questionable content in the soundtrack modality

| Label | Definition |
|---|---|
| Mature Humor | The appearance of humor involving sexual references in the video |
| Slapstick Humor | The appearance of humor involving mild physical or cartoon violence. Should be viewed as the intent of the video creator to produce comedy using violent elements |
| Gory Humor | The appearance of humor involving intense physical violence in the video |
| Sarcasm | The appearance of Passive-aggressive humorous action at the expense of another person or situation |
| Fear Horror | The appearance of things such as ghost/ dead body; creepy and crawly things; disfigurement; scary places (graveyards, old houses, overgrown forests, dungeons, attics, basements) |
| Gambling | Depictions of persons engaging in activities related to gambling, or gambling-related items |
| Foul Language | Profanities displayed in written form in the video |
| Hate speech | Hate speech displayed in written form in the video |
| Partial Nudity | Brief and/or mild depictions of nudity |
| Full Nudity | Depicting naked humans |
| Sexual Themes | Visual references to sexuality through objects or actions in a scene (but not including the sex) |
| Implicit Sexual Content | Depictions of sexual behavior, possibly including partial nudity. It implies the act of sex without showing the act itself. |
| Explicit Sexual Content | Explicit or frequent depictions of sexual behavior, possibly including nudity |
| Alcohol | Images of alcoholic beverages or consuming alcoholic beverages |
| Drug | Depiction of drug use or image of drug |
| Tobacco | Images of tobacco objects or depiction of tobacco use |
| Cartoon Violence | Violent actions involving cartoon-like situations and characters. May include violence where a character is unharmed after the action has been inflicted |
| Intense Violence | Graphic and realistic-looking depictions of physical conflict. May involve extreme and/or realistic blood, weapons, and depictions of human injury and death |
| Mild Violence | Depiction of Mild violence such as the depiction of a gun without injury or slapping and punching with no intense outcome. |
| Sexual Violence | Depiction of rape or other violent sexual acts |
| Fantasy Violence | Violent actions of a fantasy nature, involving human or non-human characters in situations easily distinguishable from rea[ST3] [SM4] l life |
| Gore | Gore is generally the aftermath and result of intense violence. This includes but is not limited to; spilled organs, dislocated and/or dismembered body parts, and blood. |
| Blood | Depiction of blood |

Table 4. Labels of objectionable content in the video modality

The idea is to design a binary annotation task. If an annotator finds the presence of content defined in the tables above, they must indicate so by selecting the specific labels. But we can also consider to use some sort of numeric scale indicating the severity of the content. This will allow us to model the different tolerance levels or thresholds we may naturally observe in different people.

Relevant Context for the Annotation Task
This section summarizes the discussion around possible content to annotate in addition to what has been depicted in Tables 1-4 above. This additional content can be considered context and there are several types of context that may be relevant for this task. Here we consider three types of context for the videos being annotated: viewer-related context, scene related context, and participants context.

Viewer related context can include the viewing history from the viewers and the viewer comments. Viewer history refers to the videos that a current user has watched prior to arriving at the current video and can also include information about who else has seen the video. We can also collect viewer comments or reactions. For example, for some cues like humor, we can capture this from comments (in some comments, it is mentioned which part of the movie is very funny). Similarly, strongly negative comments might be signals of specific types of questionable content in a video. In general, the comments themselves might represent toxic content that some users might prefer not to see or have their children have access to those comments.
The actual comments about the video and statistics such as how many people are talking about this video on social media, re-tweeted it, commented about it, or embedded it in their website. What are the analyses that can be extracted from these social sources? What is the interaction of the community regarding this specific video? Is it discussed as something positive or negative? Collecting this information can result in enabling collateral research questions.

Another type of context is related to the context of the scene being annotated. In this case, if we want to include the context, we need to provide the original video, or we can expand each scene by adding video content before and after, and in this case, we can still randomize the scenes to be annotated. In the annotation, context can be captured by adding a layer for the annotation to document these types of context.

Participant context refers to background information about people or characters appearing in the video, as well as some description of what is happening. For example, certain material is not considered appropriate communication between an adult and a minor and knowing that the participants include a minor and an adult will be relevant. Some aspects, such as sarcasm, are not necessarily bad by default. However, sarcasm can be bad in the context of aggression. In addition, background information about the participants and the events taking place can also help identify questionable content, for example, biases towards certain demographic groups, such as what kind of activity the participants are engaging in, or the gender of the participants. This additional information may help detect specific types of questionable content. Other labels that distinguish youth radicalization, the portrayal of stereotypes, or the distinction of different kinds of violence, terrorism, or political violence, as a sub-category of violence are relevant. In sum, the more comprehensive the set of labels considered, the better.

Additional Considerations for the Annotation Process
With respect to the annotation process, it is best if the annotation can be spread across a large set of annotators, as opposed to having a few numbers of them annotating most of the content. The reason for this is due to the possible negative effects of watching a lot of videos with questionable content. Perhaps some of the human annotators can be from an education background.

We might also consider asking annotators to write a short justification for labeling. This may become useful in the process of understanding biases in the annotation, but also in the process of analyzing model performance. To study the possible inherent biases in annotators, we could calculate an inter-annotator agreement that considers the intrasubject sensitivity. In other words, we use the data to calculate valid annotator-dependent normalizations of the annotations. The disadvantage of this approach is that it will demand many resources, a large number of annotations, and possibly annotating the same content multiple times. But this solution eliminates the need to create a questionnaire to discover annotator perceptions and biases towards specific types of questionable content.

The repository itself is likely to be dynamic and will evolve with time and with intended audience. For example, multiple versions of the repository can address the needs of different audiences. However, a comprehensive annotation can still be helpful and accommodate multiple communities of users, where some users can ignore certain labels and pay attention to only the labels of their choice.

One last consideration regarding the annotation process involves the annotation tool. The type of annotations we intend to collect warrant a customized annotation tool. Ideally this tool would be an online and interactive annotation framework that will make the task easy for annotators, while at the same time will provide us with quality control functionality.

## C.  Relevance of this Repository to other Areas

Researchers in ethics in AI, integrity, and biases can leverage this repository. Even though it is targeting young viewers, it can still be valid for other group demographics such as underrepresented groups. Integrity is a field where people look into the quality and the ethical implications of the content that is being propagated or shared. Having a clear set of dimensions related to ethics and/or integrity could help bootstrap, or we can use them as seeds toward other ethical questions. Some labels relevant for integrity are hate speech, fake news, pornography, gory content, the propagation of stereotypes, violence, and bullying. Responsible AI and integrity are huge organizations within Facebook as a social media company, they also exist in other companies such as Google and Amazon. What this means is that it is not a discussion taking place only in academia, but it is also an important subject on products.

## D. How to Maximize the Potential Impact of the Resources

A well-known approach to increase the visibility of the repository is to host shared tasks. These shared tasks typically consist of the lead research team releasing train/dev data and establishing a test phase for participants to submit their system predictions for ranking. We can also consider another type of shared task where we provide access to a working system built with the repository. Then, you have people trying to break the system to discover vulnerabilities of the system and then propose solutions to fix it.

Another possibility to encourage community engagement is to issue a call for shared tasks where the community proposes to us which shared task they would like to see. The lead team selects one proposal that fits well within the scope of the infrastructure and that is feasible to complete within a reasonable time span. We then design and perform the annotation task to run the shared task selected. The fact that the community provides input in the form of shared tasks proposals may result in greater participation.

## E. Discussion
Workshop participants are enthusiastic about seeing plans for developing this multimodal resource. They see many potential benefits to AI research communities, as well as a critical societal benefit in enabling the design and building of the intended technology. The main points discussed above provide a good set of recommendations that the lead research team will follow when creating the repository.
It is expected that as we work on describing and labeling questionable content, we enable the identification of pro-social content. Therefore, labeling this content will be a natural next step to take.

**Acknowledgment:** This work was supported in part by the National Science Foundation under Grant No. IIS-2036368. Any opinions, findings, and conclusions or recommendations expressed in this material are supported by the NSF award and do not necessarily reflect the National Science Foundation's views.